\documentclass{article}

    \PassOptionsToPackage{numbers, compress}{natbib}

\usepackage[preprint]{neurips_2024}

\usepackage[utf8]{inputenc} 
\usepackage[T1]{fontenc}    
\usepackage{hyperref}       
\usepackage{url}            
\usepackage{booktabs}       
\usepackage{amsfonts}       
\usepackage{nicefrac}       
\usepackage{microtype}      
\usepackage{xcolor}         
\usepackage{graphicx} 
\usepackage{wrapfig}
\usepackage{booktabs}
\usepackage{amsmath} 
\usepackage{enumitem}
\usepackage{float}
\usepackage{array}
\usepackage{multirow} 
\usepackage{graphicx}
\usepackage{amsmath}
\usepackage{multirow}
\usepackage{caption}

\title{SG-Adapter: Enhancing Text-to-Image Generation with Scene Graph Guidance}

\author{Guibao Shen$^{1,\ast}$, Luozhou Wang$^{1,\ast}$, Jiantao Lin$^1$, Wenhang Ge$^1$, Chaozhe Zhang$^1$, Xin Tao$^3$,\\
\textbf{Yuan Zhang$^3$, Pengfei Wan$^3$, Zhongyuan Wang$^3$, Guangyong Chen$^{4,5}$, Yijun Li$^6$, Ying-Cong Chen$^{1,2,\dagger}$}\\
$^1$HKUST(GZ), $^2$HKUST, $^3$Kuaishou, $^4$Zhejiang Lab, $^5$Zhejiang University, $^6$Adobe Research
}

\begin{document}

\maketitle

\renewcommand{\thefootnote}{}
\footnotetext{$^\ast$Equal contribution}
\renewcommand{\thefootnote}{}
\footnotetext{$^\dagger$Corresponding author}

\begin{abstract}
  Recent advancements in text-to-image generation have been propelled by the development of diffusion models and multi-modality learning. However, since text is typically represented sequentially in these models, it often falls short in providing accurate contextualization and structural control. So the generated images do not consistently align with human expectations, especially in complex scenarios involving multiple objects and relationships. In this paper, we introduce the \textbf{Scene Graph Adapter} (SG-Adapter), leveraging the structured representation of scene graphs to rectify inaccuracies in the original text embeddings. The SG-Adapter's explicit and non-fully connected graph representation greatly improves the fully connected, transformer-based text representations. This enhancement is particularly notable in maintaining precise correspondence in scenarios involving multiple relationships. To address the challenges posed by low-quality annotated datasets like Visual Genome~\cite{krishna2017visual}, we have manually curated a highly clean, multi-relational scene graph-image paired dataset MultiRels. Furthermore, we design three metrics derived from GPT-4V\cite{achiam2023gpt} to effectively and thoroughly measure the correspondence between images and scene graphs. Both qualitative and quantitative results validate the efficacy of our approach in controlling the correspondence in multiple relationships.
\end{abstract}

\section{Introduction}
\vspace{-0.1in}
\label{sec:intro}

Image generation has made great progress thanks to the success of a series of text-to-image diffusion models \cite{rombach2022high, saharia2022photorealistic, ramesh2022hierarchical, ho2020denoising, nichol2021glide, dhariwal2021diffusion,song2020denoising}. Due to the huge amount of text-image paired training data\cite{schuhmann2022laion,lin2014microsoft} and the numerous model parameters, these text-conditioned models show fantastic image quality and attract great attention from researchers as well as industries.

\begin{figure}[t]
    \centering
    \includegraphics[width=1.0\linewidth]{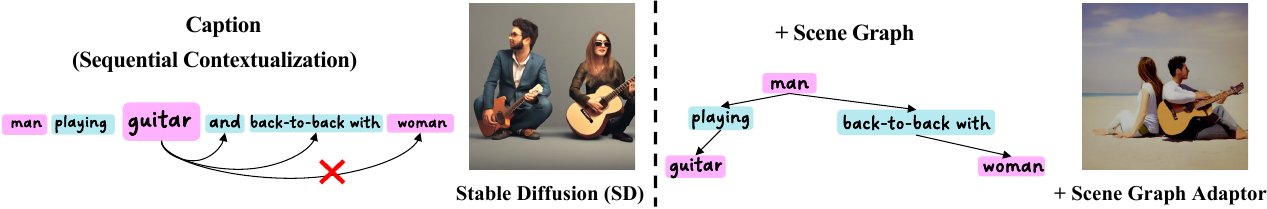}
    \vspace{-0.25in}
    \caption{\textbf{Overcoming Contextualization Limits in Image Generation with Scene Graph}. The left section highlights the limitations of text embeddings in sequential text processing, showcasing how relations like ``playing guitar'' may erroneously apply to the ``woman''. The right section illustrates the improvements of using a Scene Graph, which provides structured clarity, enabling precise relation. 
    }
    \label{fig-intro}
    \vspace{-0.25in}
\end{figure}

However, the text encoder used by these models are suboptimal for generating coherent images. They often face challenges in contextualizing text tokens, which requires capturing both their intrinsic meanings and their contextual understanding from surrounding tokens. 
The most commonly used text encoder, the CLIP model\cite{radford2021clip}, applies causal attention mechanisms to allow tokens to gather information from preceding ones.
However, this sequential processing can lead to the ``leakage issue'', where relations or attributes incorrectly influence unrelated elements in the image.
For example, in ``A man playing the guitar back to back with a woman'', the relation ``playing guitar'' might be mistakenly applied to the woman due to the linear text sequence (Fig.~\ref{fig-intro}). Such cases, where the encoder fails to recognize distinct entity boundaries, underscore the need for more advanced text embedding methods capable of accurately depicting separate entities and their respective relations in complex scenes. 

Scene graphs, with their structured representation, effectively avoid the contextualization issues inherent in sequential text tokens. 
Scene graphs depict images as networks of entities (nodes) and their relationships (edges), ensuring clear, non-linear associations. Relations are directly linked to specific nodes, preventing the ambiguity and misinterpretation common in text-based processing. 
This distinct structure allows for precise and unambiguous representation of complex scenes, enhancing the accuracy of image generation.
This capability positions them as a natural control signal for image generation. 
However, pure scene graph-to-image generation \cite{yang2022diffusion, johnson2018sg2i, farshad2023scenegenie, li2019pastegan} lags behind text-to-image generation in terms of image quality and is far from practical use. This is because the scene graph-image pair data are considerably smaller than text-image pair data. For instance, Visual Genome\cite{krishna2017visual} includes 108,077 pairs with limited-quality scene graph labels. In stark contrast, the typical text-image dataset LAION-5B\cite{tang2020unbiased} contains a massive 5.85 billion CLIP-filtered image-text pairs. 

Acknowledging the challenges in text-to-image generation and scene graph datasets, our research aims to leverage the limited scene graph annotations to enhance the control and accuracy of text-to-image generation. 
Recent studies on adapters \cite{li2023gligen, zhang2023controlnet, mou2023t2i}, in pre-trained text-to-image models offer promising directions. 
These adapters introduce new control mechanisms that retain the model's image generation quality. 

In this study, we re-examine the text-to-image generation process, pinpointing issues of incorrect contextualization arising during text encoder computation. 
This problem is traced back to the causal attention mask, which is used in the pre-trained CLIP text model \cite{radford2021clip}. 
It fails to account for the structural semantics inherent in captions.
However, directly replacing this causal attention mask during inference is ineffective through our observations, as presented in Appendix.\ref{subsec:init_attempt}.
Consequently, we propose the Scene Graph Adapter (SG-Adapter), strategically plugged in after the CLIP text encoder, to tackle this issue.
This adapter utilizes scene graph knowledge to refine text embeddings. 
Critically, to ensure precise contextualization, we design our adapter based on a transformer architecture and incorporate a novel triplet-token attention mask called Scene Graph(SG) Mask, fostering accurate and contextually relevant text-to-image generation.

On the other hand, the quality of scene graph annotations often suffers due to various issues like limited rationality, language biases, or reporting biases. 
As a result, many visual relationships in these annotations tend to be simplistic and less informative \cite{tang2020unbiased}.
Traditional datasets are inadequate for experimental observation. 
To circumvent these limitations, we conducted preliminary experiments on a smaller-scale but higher-quality dataset Reversion\cite{huang2023reversion}. 
This approach demonstrated the effectiveness of our method. 
Subsequently, we expanded our focus to contribute a more complex, multi-relational dataset with highly clean annotations called MultiRels that better encapsulate complex semantic structures. 
Our SG-Adapter, finetune on MultiRels, shows a strong ability to generate relations with precise correspondences. 
Moreover, considering traditional evaluation metrics like FID and CLIP-Score cannot perceive the abstract Relation Correspondence concept, we propose three evaluation metrics derived from the advanced GPT-4V\cite{achiam2023gpt} to evaluate the correspondences between images and scene graphs.
In summary, our contributions are as follows:
\begin{itemize}[leftmargin=*, noitemsep, nolistsep]
\item We propose our SG-Adapter to correct the incorrect contextualization in text embeddings that results in ``relation leakage''. Our adapter effectively addresses this issue and enhances the structural semantics generation capabilities of current text-to-image models.
\item We expand the ReVersion dataset to include multiple relations, creating a more comprehensive dataset with high-quality annotations called MultiRels that could be adopted to better demonstrate the generation models' ability to handle complex structural semantics. Besides, for effective and fair comparison in terms of relation accuracy, we contribute three metrics derived from GPT-4V\cite{achiam2023gpt}.
\item Both qualitative and quantitative results illustrate that our SG-Adapter outperforms state-of-the-art text-to-image and SG-to-image methods in terms of relation generation and control.
\end{itemize}


\section{Related Work}
\label{sec:related_work}

\noindent\textbf{Text-to-Image Generation.}~The evolution of text-to-image diffusion models \cite{rombach2022high, saharia2022photorealistic, ramesh2022hierarchical, ho2020denoising, nichol2021glide, dhariwal2021diffusion, song2020denoising} has marked a significant advancement in the generation of high-quality images. Among these, Stable Diffusion (SD) \cite{rombach2022high} is particularly notable, utilizing a pre-trained autoencoder alongside a diffusion model to effectively transform textual prompts into precise and detailed visual representations. Despite their impressive capabilities, these models still face challenges, particularly in terms of the leakage issue, where relations or attributes inadvertently influence unrelated elements within an image \cite{feng2022ssd,li2023divide}. While \cite{li2023divide} addresses this through a novel Jensen-Shannon divergence-based binding loss, it does not fully acknowledge the issue as stemming from the improper contextualization of text embedding. In a similar vein, \cite{feng2022ssd} employs linguistic structures such as constituency trees or scene graphs to guide the diffusion process. However, this method does not entirely address the root cause of these issues but constructs a set of text embeddings and only partially fixes the wrong contextualization, which is linked to the causal attention mechanism used in the CLIP text encoder.

\noindent\textbf{Scene Graph-to-Image Generation.}~In the realm of scene graph to image generation, some works often rely on scene layouts as an intermediate representation \cite{johnson2018sg2i, farshad2023scenegenie} in a two-stage pipeline. They first create scene layouts from scene graphs and then generate images based on the layouts. These layouts, serving as image-like representations of scene graphs, can often be suboptimal due to manual crafting and potential misalignment with actual scene graphs. In this paper, we clarify that \textbf{Layout-to-Image\cite{zheng2023layoutdiffusion, li2023gligen, chen2023trainingfreelayout, couairon2023zerolayout} and Scene Graph-to-Image are separate different topics}, as demonstrated in Fig.\ref{fig:relation_layout}. Layout-to-Image aims to generate objects in user-specified areas while having limited ability to represent complicated Relations. This limitation has led to the development of alternative approaches, such as SGDiff \cite{yang2022sgdiff}, which optimizes scene graph embeddings for better alignment with images, and other works \cite{wu2023sggan} that introduce ``knowledge consensus'' as a means to disentangle complex semantics between knowledge graphs and images. Despite the advances in this field, the dearth of large-scale and high-quality data continues to impede its widespread practical application.

\noindent\textbf{Single Relation Learning.} Though Text-to-Image generation models have made great progress, they might fall short when generating complicated single relationships. RRNet\cite{wu2024relation} learns a relationship via a heterogeneous GCN while Reversion\cite{huang2023reversion} reverses the target relation from example images by optimizing the relation prompt. Instead of addressing the issue of single relations, our study focuses on generating multiple relations with accurate correspondences.

\noindent\textbf{Adapter.}~The integration of adapters into existing models has emerged as a noteworthy innovation, exemplified by methods such as ControlNet \cite{zhang2023controlnet} and GLIGEN \cite{li2023gligen}. These techniques have enabled models like SD to enhance user control without requiring extensive retraining. For example, ControlNet \cite{zhang2023controlnet} operates as a plug-in module, extracting and integrating residual features from each image condition into SD's U-Net for enhanced control. GLIGEN \cite{li2023gligen}, on the other hand, incorporates location control by integrating a new self-attention module into the U-Net.

Our work critically examines the issue of false contextualization in SD and develops a scene graph-only trained adapter to enhance the semantic structure controllability of SD, addressing a gap in the current landscape of text-to-image generation.

\section{Proposed Method}

In this work, we aim to rectify the incorrect contextualization in text embedding caused by the prevalent causal attention masks in language models, \textit{i.e.}, CLIP text encoder \( E_T(\cdot) \).

\subsection{Discussions of Causal Attention}

The causal attention mask in the context of transformer models ensures that the prediction for a specific token can only consider previously generated tokens. 
For a caption \(c\) of \(N\) tokens, the causal attention mask \(\mathbf{M}\) is an \(N \times N\) matrix where each entry \(M_{ij}^{\text{causal}}\) is defined as:

\begin{equation}
\mathbf{M}_{ij}^{\text{causal}} = 
\begin{cases} 
0, & \text{if } j \leq i, \\
-\infty, & \text{if } j > i. 
\end{cases}
\label{causal-text-attention}
\end{equation}

This matrix is utilized in the scaled dot-product attention mechanism within the transformer model. The attention scores with the mask are computed as follows:

\begin{equation}
\text{Attention}(\mathbf{Q}, \mathbf{K}, \mathbf{V}, \mathbf{M}) = \text{softmax}\left(\frac{\mathbf{Q}\mathbf{K}^T}{\sqrt{d_k}} + \mathbf{M}^{\text{}}\right)\mathbf{V},
\label{attention}
\end{equation}
where \( \mathbf{Q} \), \( \mathbf{K} \), and \( \mathbf{V} \) denote the query, key, and value matrices respectively, and \( d_k \) represent the dimension of the key vectors that are used for scaling. 
The softmax function is applied subsequent to the addition of the causal attention mask \( \mathbf{M}^{\text{causal}} \). 
This ensures that for each position \( i \) in the sequence, the attention is only on positions \( j \leq i \), effectively masking out the future tokens by assigning them a value that approximates zero probability in the softmax due to the \( -\infty \) entries in the mask.

\begin{figure*}[t]
    \centering
    \includegraphics[width=1.0\linewidth]{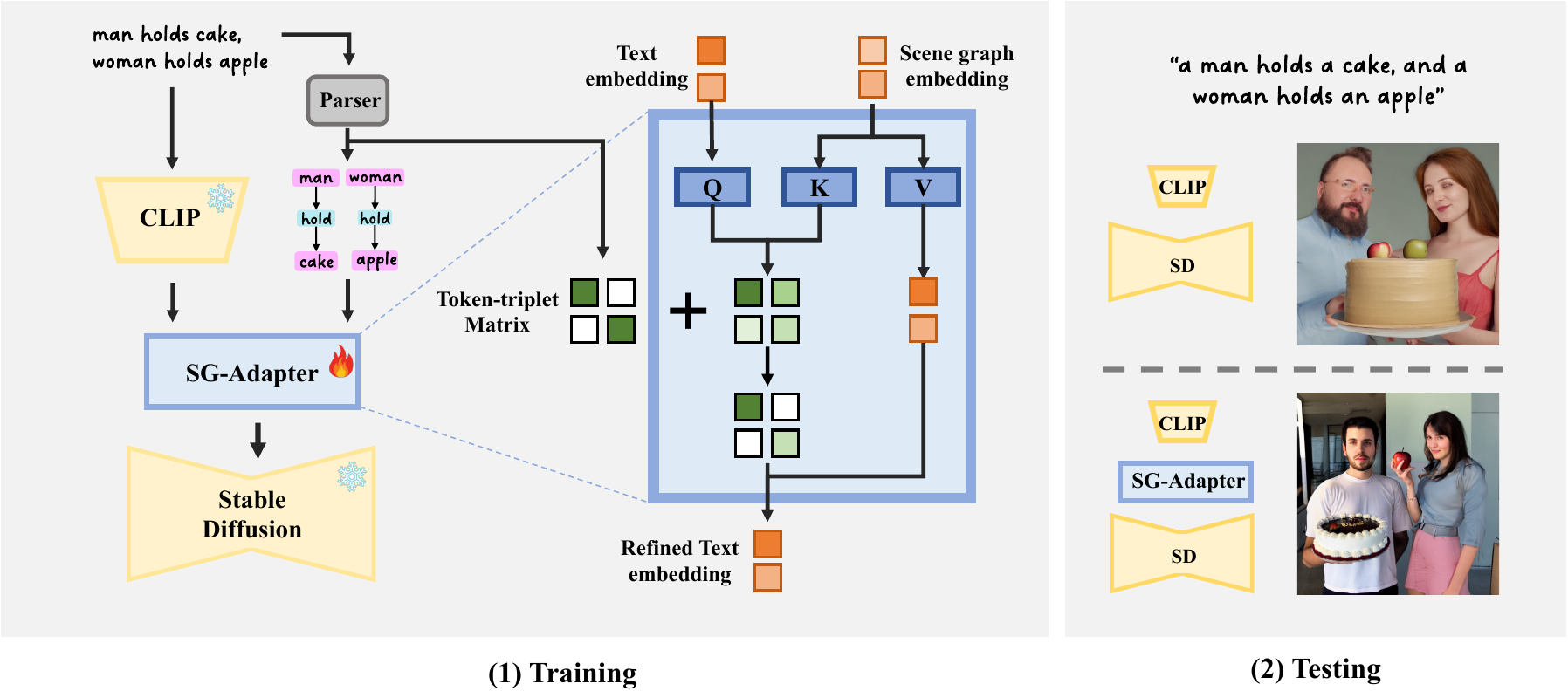}
    \vspace{-0.2in}
    \caption{\textbf{Framework for SG-Adapter in Stable Diffusion.} The \textbf{Parser} (could be either an NLP tool\cite{feng2022ssd} or GPT-4\cite{achiam2023gpt}) extracts linguistic structures from text inputs. \textbf{Scene graph embeddings} are computed as per Eq~\eqref{sg-embedding}. The \textbf{token-triplet matrix}, generated by the function \( \tau \), guides the refinement of each token and its associated triplet. During testing, when integrated with our SG-Adapter, Stable Diffusion more accurately captures the intended semantic structure in the generated images.}
    \label{fig-method-pipeline}
\end{figure*}

However, the causal attention mask may fail to maintain the integrity of subject-relation-object bindings within captions, leading to incorrect contextual associations between entities and relations.
%
For instance in Fig.~\ref{fig-method-pipeline}, the caption ``a man holds a cake and a woman holds an apple'' contains two distinct subject-relation-object bindings. 
However, traditional causal attention will not mask out the former token ``holds a cake'' when computing the embedding of ``a woman'' and may incorrectly leak the semantics of ``holds a cake'' to ``a woman'', disrupting the semantic structure of the caption.


To enable precise interactions among tokens, a thorough understanding of the semantic structure encapsulated in the caption is crucial. 
This structure is aptly represented by a scene graph, constituted of a sequence of triplets formatted as \(\langle\text{subject, relation, object}\rangle\). 
The extraction of these \(K\) triplets from the caption can be achieved either through an NLP parser \cite{feng2022ssd} or GPT-4\cite{achiam2023gpt}. 
Each triplet \( \mathcal{T}_k \) is indexed by \( k \), and a token-to-triplet mapping function \( \tau(\cdot) \) is defined, which maps each token \( i \) to its respective triplet index. The prompt used in GPT-4 is provided in Appendix.~\ref{subsec:prompt}.

Our analysis stipulates that interaction between any two tokens \( i \) and \( j \) is permissible solely when they are constituents of an identical triplet, as determined by the condition \( \tau(i) = \tau(j) \).
In light of this, we propose refining the causal attention mask such that it facilitates the alignment of each token with its associated triplet. 
This refinement process is intended to bolster the model's ability to discern and maintain contextual relationships.
The attention mask \( \mathbf{M}^{\tau} \), adjusted for triplet alignment, is formally given by:
\begin{equation}
\mathbf{M}_{ij}^{\tau} = 
\begin{cases} 
0, & \text{if } \tau(i) = \tau(j), \\
-\infty, & \text{otherwise}.
\end{cases}
\label{sg-text-attention}
\end{equation}

With the refined attention mask, the most straightforward approach would be to replace the CLIP attention mask during inference. 
However, this can cause out-of-distribution issues, as shown in the Appendix.~\ref{subsec:init_attempt}. 
Additionally, training a CLIP model from scratch using a scene graph attention mask requires substantial resources.

\subsection{Scene Graph Guided Generation}

\vspace{0.5em}
\noindent\textbf{Scene Graph Representation.}
To address these challenges, we devised an alternative approach that augments the text embeddings produced by the CLIP model with the scene graph information. 
The text embedding output by the CLIP model is denoted as \( \mathbf{w} = E_T(c) \), where \( \mathbf{w} \in \mathbb{R}^{N \times D} \) and \( D \) signifies the dimensionality of the embedding vectors by CLIP Text Encoder \( E_{T} \). Here, \( \mathbf{w}_i \in \mathbb{R}^{D} \) represents the embedding of the \( i^{th} \) token.

To encapsulate the scene graph information, we construct a unified embedding for each semantic triplet. For a given triplet \( \mathcal{T}_k = \langle s_k, r_k, o_k \rangle \), where \( s_k \), \( r_k \), and \( o_k \) denote the subject, relation, and object of the triplet respectively, we apply \( E_{T} \) to each component to obtain their embeddings. These embeddings are then concatenated to form a composite triplet embedding as follows:

\begin{equation}
\mathbf{e}_{k} = l(\text{concat}(E_{\text{T}}(s_k), E_{\text{T}}(r_k), E_{\text{T}}(o_k))),
\label{sg-embedding}
\end{equation}
where \( l(\cdot) \) is a projection function that maps its input to the dimensionality \( D \). 
The final scene graph embedding \( \mathbf{e} \in \mathbb{R}^{K \times D} \) is the concatenation of all the triplet embeddings. 
This embedding is then utilized to refine the embeddings of corresponding tokens.

\vspace{0.5em}
\noindent\textbf{Scene Graph Adapter.}
To implement the refinement, an adapter \( {f}(\cdot) \) is designed. 
Conceptualized as a transformer module, our adapter predominantly integrates a cross-attention layer. 
This layer facilitates relation-text attention by leveraging the scene graph embedding \( \mathbf{e} \) to calculate the keys \( \mathbf{K} \) and values \( \mathbf{V} \), and the text embedding \( \mathbf{w} \) to calculate the queries \( \mathbf{Q} \). 
Such interaction is pivotal in updating the text embeddings with relation-specific nuances. The model \( f \) is mathematically represented as:

\begin{equation}
\begin{aligned}
\mathbf{w}' &= f(\mathbf{w}, \mathbf{e}, \mathbf{M}) \\
&= \text{Attention}(\mathbf{Q}, \mathbf{K}, \mathbf{V}, \mathbf{M}^{\textbf{sg}}),
\end{aligned}
\label{sg-adapter}
\end{equation}
where \( \mathbf{Q} =l_Q(\mathbf{w}) \), \( \mathbf{K} =l_K(\mathbf{e}) \), and \( \mathbf{V} =l_V(\mathbf{e}) \) are derived using respective projection layers in the attention framework. The term \( \mathbf{w}' \) represents the improved text embeddings resulting from the application of the cross-attention mechanism.

To ensure that each token embedding \( \mathbf{w}_i \) attends to the appropriate triplet embedding \( \mathbf{e}_{\tau(i)} \), we introduce the token-triplet attention mask \( \mathbf{M}^{\textbf{sg}} \in \mathbb{R}^{N \times K} \). This mask is designed to allow each token embedding to attend exclusively to its corresponding triplet embedding. The mask is formalized as follows:

\begin{equation}
\mathbf{M}_{ik}^{\textbf{sg}} = 
\begin{cases} 
0, & \text{if } \tau(i) = k \\
-\infty, & \text{otherwise}
\end{cases}
\label{sg-attention}
\end{equation}
and implementing this mask within the cross-attention layer of our model permits \( \mathbf{w}_i \) to be selectively refined based on the contextual relevance of \( \mathcal{T}_{\tau(i)} \). Such a targeted approach ensures precise contextualization within the generated scene graph.

To train our model, we leverage the pre-trained diffusion model framework. Given an image \( x \), its corresponding text embedding \( \mathbf{w} \), and the extracted scene graph embedding \( \mathbf{e} \), our model undergoes a training phase analogous to that of the diffusion model. The objective is to minimize the discrepancy between the predicted and actual noise variables. The training loss function is defined as:

\begin{equation}
\mathcal{L}_t=\mathbb{E}_{\mathbf{x}, t, \epsilon}\left[ \left\Vert \epsilon_t - \epsilon_{\theta}(\mathbf{x_t}, t, f(\mathbf{w}, \mathbf{e}, \mathbf{M}^{sg})) \right\Vert^2_2\right],
\label{sg-adapter-train}
\end{equation}
where \( \epsilon_t \) represents the noise variable at time step \( t \), and \( \epsilon_{\theta} \) denotes the noise prediction from the pre-trained diffusion model parameterized by \( \theta \). The function \( f(\mathbf{w}, \mathbf{e}, M) \) encapsulates the adapter's output, which refines the text embedding with relation-specific context derived from the scene graph embedding. 
Our training algorithm iteratively adjusts the parameters of \( f(\cdot ) \) to reduce \( \mathcal{L}_t \), thereby improving the fidelity of generated images to the input descriptions and scene graph structure.

\section{Experiments}
\label{sec:experiments}


\subsection{Dataset Configuration}
\label{subsec:dataset}
As we mentioned above, a scene graph dataset with clean and precise relational annotations is essential as well as effective for learning relation representation. 
To train a model that can encode both single and multiple relations accurately, for the first time, we present a multiple relations scene graph-image paired dataset with highly precise labels called \textbf{MultiRels}. 
The dataset with size 309 is composed of two parts as follows:
\begin{itemize}[leftmargin=*, noitemsep, nolistsep]
    \item ReVersion \cite{huang2023reversion}: This part consists of 99 images with only a single clear relation. There are 10 representative relations in Reversion and most of them are ``difficult'' relations that current text-to-image models can not generate well, e.g., \textit{shake hands with}, \textit{sit back to back}, \textit{is painted on}. Reversion is mainly used to learn new single relations.
    \item Multiple Relations: For multi-relations learning, in this part, we manually collect images with 1-4 (mostly more than 1) salient relations and label them with accurate scene graphs. Most of the relations are ``simple'' relations (e.g., \textit{holding}, \textit{drinking}, \textit{stands on}, \textit{sits on}) that current text-to-image models generate well individually but fail when existing multiple objects and multiple relations. There are also small parts of relations that are ``difficult'' like \textit{is above}, \textit{is under}. Multiple Relations is towards to train a model that could generate multiple relations with correct correspondences.
\end{itemize}
Besides the scene graph, we additionally provide a token-triplet matrix for each image. 

\vspace{0.5em}
\noindent\textbf{Test Scenarios.} To fairly validate the effectiveness and generalization performance of our method, we design 20 different testing scenarios that regroup 2-3 randomly selected relations in MultiRels. 
For more details about the MultiRels, please refer to our Appendix.

\subsection{Baseline Methods}
\label{subsec:adapter_comp}
\begin{figure*}[!htbp]
    \centering
    \includegraphics[width=1.0\linewidth]{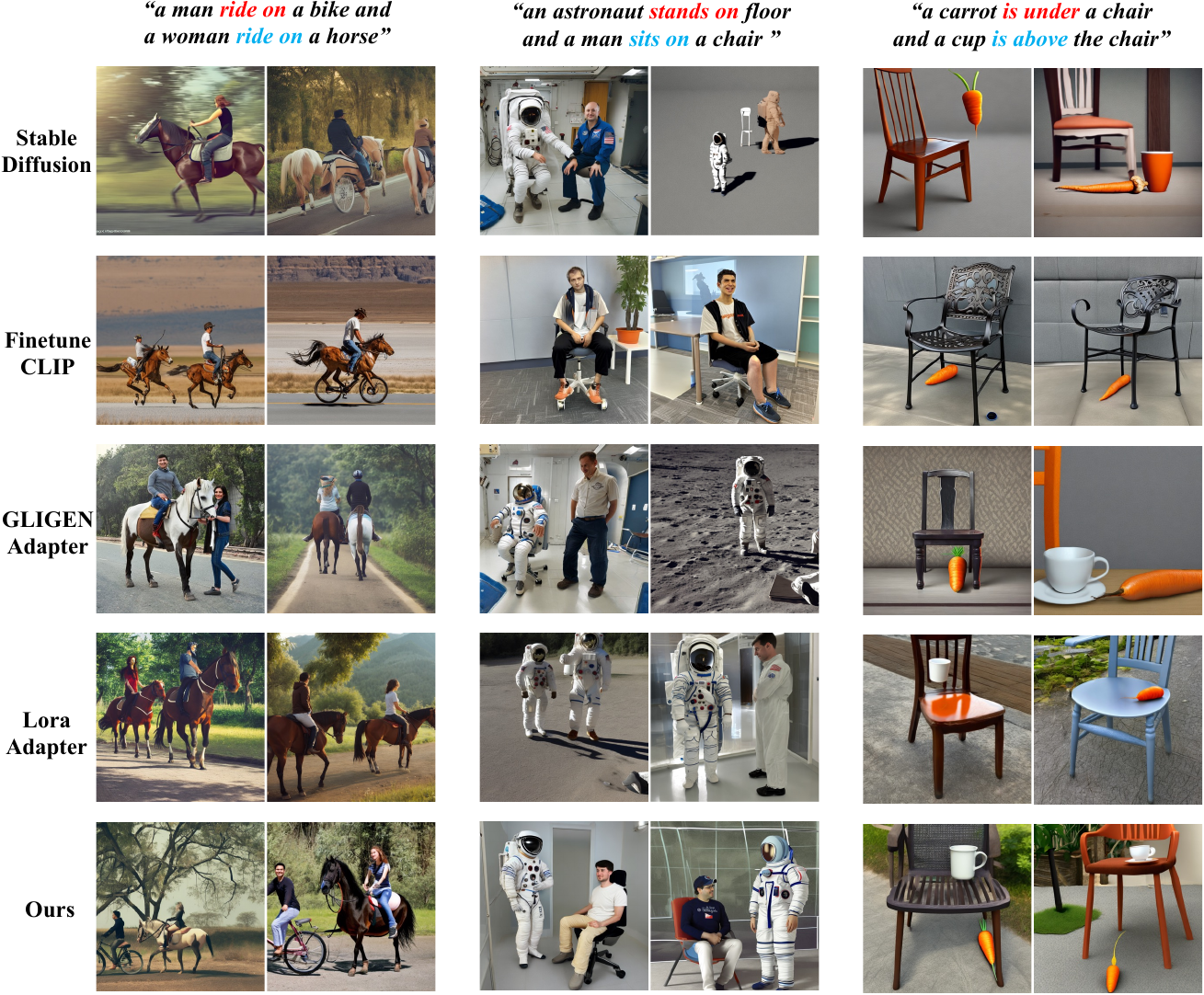}
    \vspace{-0.1in}
    \caption{\textbf{Qualitative Comparisons} with Adaptation Methods. In addition to precisely generating each individual relation in the text prompt, our SG-Adapter successfully creates all multiple relations together in correct correspondence.}
    \label{fig-comparison-all}
    \vspace{-0.2in}
\end{figure*}

In our experiments, we evaluate the performance of SG-Adapter against several alternative methods, each with a distinct approach to text-to-image generation:

\begin{itemize}[leftmargin=*, noitemsep, nolistsep]
\item \textbf{Fine-tuning (FT):} No adapters are used. We directly fine-tune components of Stable Diffusion, CLIP text encoder according to our analysis in Sec.\ref{sec:intro}.
\item \textbf{LoRA Adapter:} Utilizes the LoRA (\textit{Low-Rank Adaptation}) adapter \cite{hu2021lora}. The LoRA adapter does not integrate direct scene graph information in the adaptation process.
\item \textbf{Gated Self-Attention Adapter:} Employs the adapter proposed by GLIGEN \cite{li2023gligen} which introduces scene graph tokens into the model. For generating scene graph tokens, we use a specialized scene graph encoder that has been pre-trained for this specific task \cite{yang2022sgdiff}.
\end{itemize}

Note these baseline methods also assess the impact of Scene Graph Embedding from two critical angles: \begin{itemize}[leftmargin=*, noitemsep, nolistsep]
\item \textbf{Inclusion of SG Embed}: We explore the necessity of SG Embed by contrasting the performance of our SG-Adapter with the LoRA Adapter, which operates without SG Embed. 
\item \textbf{Placement of SG Embed}: We assess the effect of embedding placement by comparing our model to the GLIGEN Adapter, which integrates SG Embed within a U-Net architecture.
\end{itemize}

\subsection{Qualitative Evaluation} We begin with a qualitative assessment, showcasing synthetic images from each method. SG-Adapter's ability to accurately depict complex relational structures and entities is highlighted, with Fig.~\ref{fig-comparison-all} illustrating its superiority in maintaining correct relation correspondence in generated images. More qualitative results are provided in our Appendix.

\subsection{Quantitative Evaluation}

\begin{table}[h]
\centering
\caption{\textbf{Quantitative Evaluation} of each method in terms of Automatic Relational Metrics, Human Evaluations, and Image Quality.}
\renewcommand{\arraystretch}{1.2} 
\scriptsize 
\resizebox{1.0\columnwidth}{!}{%
\begin{tabular}{c c c c c c c}
\hline
\multirow{3}{*}{\textbf{Method}} & \multicolumn{3}{c}{\textbf{Automatic Metrics}} & \multicolumn{2}{c}{\textbf{Human Evaluations}} & \multirow{3}{*}{\textbf{FID~$\downarrow$}} \\ \cline{2-6} 
                                 & \textbf{SG-} & \textbf{Entity-} & \textbf{Relation-} & \textbf{Relation-} & \textbf{Entity-} &   \\ 
                 & \textbf{IoU~$\uparrow$} & \textbf{IoU~$\uparrow$} & \textbf{IoU~$\uparrow$} & \textbf{Accuracy~$\uparrow$} & \textbf{Accuracy~$\uparrow$} &  \\ \hline
Stable Diffusion~\cite{rombach2022high} & 0.157 & 0.673 & 0.526 & 5.38\% & 5.48\% & \textbf{25.0} \\ 
Finetune CLIP    & 0.198 & 0.499 & 0.635 & 5.38\% & 6.78\% & 58.2 \\ 
GLIGEN Adapter~\cite{li2023gligen}   & 0.141 & 0.689 & 0.546 & 5.72\% & 5.58\% & 27.4 \\ 
LoRA Adapter~\cite{hu2021lora}     & 0.145 & 0.653 & 0.540 & 5.96\% & 5.05\% & 27.5 \\ \hline
\textbf{SG-Adapter}       & \textbf{0.623} & \textbf{0.812} & \textbf{0.753} & \textbf{77.6\%} & \textbf{77.1\%} & 26.2 \\ \hline
\end{tabular}
}
\vspace{-0.2in}
\label{tab:quanti_comparison}
\end{table}

\noindent\textbf{Automatic Metrics.} Since relation correspondence in images is complicated and abstract, existing metrics like FID can not perceive such conception. To fairly and effectively evaluate the ability to generate accurate relations of each method, we contribute three metrics derived from the advanced GPT-4V\cite{achiam2023gpt}: Scene Graph(SG)-IoU, Relation-IoU, and Entity-IoU. 
These metrics are computed as follows: Given a generated image \( I \), we obtain the scene graph represented as a list of triplets: 
\(
\hat{\mathcal{T}} = \{ \hat{\mathcal{T}}_1, \hat{\mathcal{T}}_2, \ldots, \hat{\mathcal{T}}_n = \langle \hat{s}_n, \hat{r}_n, \hat{o}_n \rangle\}  
\)
and the entity list \(\mathbf{\hat{\xi}}\) using GPT-4V, denoted as \( \hat{\mathcal{T}}, \hat{\xi} = \mathrm{GPT}(I) \).The extraction prompt is provided in the appendix, see \ref{subsec:prompt}. From \(\hat{\mathcal{T}}\), we can also derive the relation list \(\mathbf{\hat{r}}\).  Then, given the input scene graph \(\mathcal{T}\) and entity list \(\mathbf{\xi}\) derived from \(\mathcal{T}\), the three metrics are computed as follows:
\[
\mathrm{SG\text{-}IoU} = \mathrm{IoU}(\mathcal{T}, \hat{\mathcal{T}}), \quad
\mathrm{Entity\text{-}IoU} = \mathrm{IoU}(\xi, \hat{\xi}), \quad
\mathrm{Relation\text{-}IoU} = \mathrm{IoU}(\mathbf{r}, \mathbf{\hat{r}})
\]

SG-IoU indicates whether each relationship is well generated according to input correspondence and the other two metrics show whether each object and relation are generated.

\noindent\textbf{Human Evalution.}
 We conduct a user study engaging \textbf{104} participants in evaluating \textbf{20} test scenarios, each comprising textual descriptions and corresponding sets of images from the evaluated methods. We collect the percentage of user preference for each method in terms of the following criteria:

\noindent \textit{Entity Accuracy}: Evaluating the precision in representing all textual entities within the images.

\noindent \textit{Relation Accuracy}: Assessing the accuracy of depicted relationships between entities, ensuring alignment with the textual descriptions.

\noindent\textbf{Image Quality.} We compute the FID\cite{heusel2018gans} which quantifies the discrepancy between the distribution of generated images and 5000 validation images from the MS-COCO-Stuff\cite{caesar2018coco}.

\noindent\textbf{Quantitative Analysis.} 
\begin{itemize}[leftmargin=*, noitemsep, nolistsep]
\item \textit{Relation Correctness.} As demonstrated in Tab.~\ref{tab:quanti_comparison}, SG-Adapter consistently outperforms the baseline methods on both Automatic Metrics and Human Evaluations, significantly on SG-IoU and Relation Accuracy, showing a strong ability to generate relations with precise correspondences. Besides, high Entity-IoU and Relation-IoU while low SG-IoU indicate that though all the baseline methods successfully generated the required entity and relation, they can not distribute them to each other with correct correspondence.
\item {Image Quality Maintenance.} Finetuning a pre-trained T2I model on a relatively small dataset will degrade the FID inevitably\cite{wang2023evaluation,ruiz2023dreambooth}. As presented in Tab.\ref{tab:quanti_comparison}, our SG-Adapter maintains the image quality best among all the alternatives. 
\end{itemize}




\subsection{Ablation Study}

\begin{table}[!t]
    \caption{Ablation study on SG Mask.}
    \centering
    \resizebox{0.7\columnwidth}{!}{%
    \begin{tabular}{ccccc}
    \toprule
        \textbf{Method} & \textbf{SG-IoU~$\uparrow$}& \textbf{Entity-IoU~$\uparrow$}& \textbf{Relation-IoU~$\uparrow$}& \textbf{FID~$\downarrow$}\\ \midrule
        \textbf{w/o SG Mask} & 0.316 & 0.742 & 0.668 & 26.7\\
        \textbf{SG-Adapter} & 0.623 & 0.812  & 0.753 & 26.1\\ \bottomrule
    \end{tabular}
    }
    \label{tab:ablation}
    \vspace{-0.2in}
\end{table}

To elucidate the contributions of SG-Adapter's distinct components, we performed a comparative analysis by systematically ablating specific features. Through our experiments, we have demonstrated that the inclusion of scene graph embeddings is essential and that integrating them into the text encoder is more effective. In this section, we conduct an ablation study on the internal elements of our method:

\noindent \textbf{Token-to-Triplet Causal Mask (SG Mask)}: We assessed the significance of the causal mask designed to map tokens to triplets, evaluating its role in refining the image generation process.

\vspace{-0.0in}
\noindent As shown in Tab.~\ref{tab:ablation}, our method performs better than the one without SG mask on all evaluation metrics, indicating the effectiveness of the proposed SG mask.

\subsection{SG-to-Image Generation Evaluation}
\label{subsec:sg2i}

We further benchmark our method against state-of-the-art SG-to-Image generation approaches including SG2IM\cite{johnson2018sg2i}, PasteGAN\cite{li2019pastegan}, SceneGenie\cite{wu2023sggan} and SGDiff\cite{yang2022sgdiff} both quantitatively and qualitatively. (Note that SceneGenie does not release the code, we just report the quantitative result from their paper.)
We train our SG-Adapter on the commonly used Scene Graph dataset~\cite{plummer2015flickr30k} to demonstrate that our method is extendable to large-scale datasets. For Flickr30k~\cite{plummer2015flickr30k}, we employed NLP processing techniques\cite{feng2022ssd} to extract scene graph information.

\begin{figure*}[htbp]
    \centering
    \includegraphics[width=1.0\linewidth]{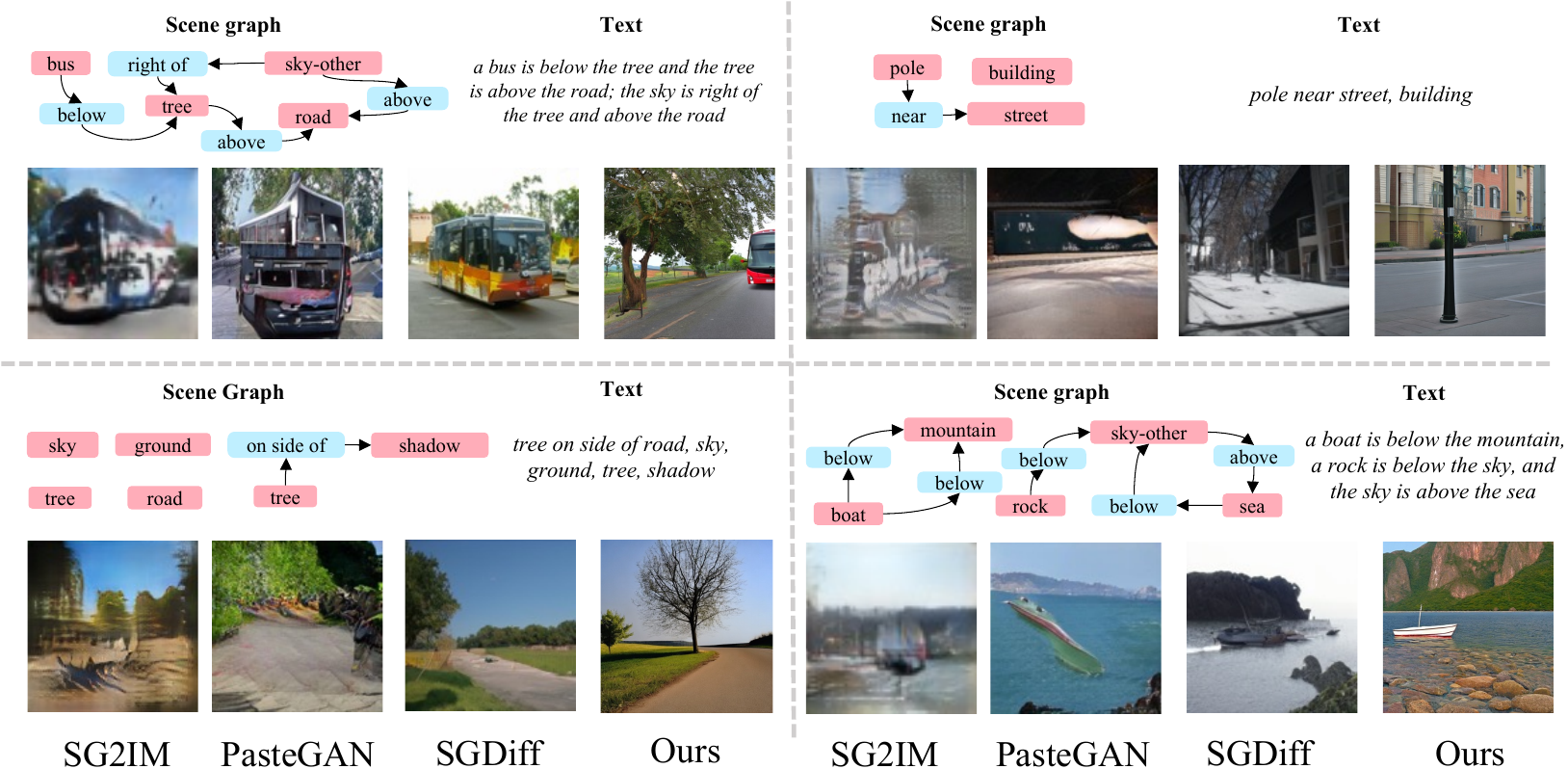}
    \caption{\textbf{Comparison with Scene Graph to Image Generation}. SG-Adapter outperforms other SG generation methods in terms of image quality and relation accuracy.}
    \label{fig-exp-sg}
    \vspace{+0.0in}
\end{figure*}

For qualitative comparison, we present generated images of each method in Fig.~\ref{fig-exp-sg}. 
Benefiting from our learned adapter strategy, we harness the high-resolution and high-fidelity generative capabilities of pre-trained text-to-image models, thereby achieving superior image quality in our results. Moreover, our method adeptly renders complex relational structures in images, signifying that our adapter effectively leverages scene graph information to accurately reflect intricate relationships within the visual output.

For quantitative comparison, due to the large size of the validation sets in the above datasets, performing GPT-4V-based evaluations for all methods would require a significant number of API calls, exceeding our available resources. 
Therefore, we primarily evaluated image quality metrics. 
We provided the FID and Inception Score~\cite{salimans2016improved} for each method in Tab.~\ref{tab:sg}. 
The results were also tested on the COCO-Stuff validation dataset with a resolution of 256$\times$256. Tab.~\ref{tab:sg} shows that SG-Adapters outperform other SG-to-Image approaches on both evaluation metrics by a significant margin, demonstrating superiority in both image fidelity and diversity.
\begin{table}[htbp]
    \caption{Comparison with SG2I Generation Methods.}
    \centering
    \resizebox{0.9\columnwidth}{!}{%
    \begin{tabular}{cccccc}
    \toprule
        \textbf{Method} & \textbf{SG2IM}~\cite{johnson2018sg2i}& \textbf{PasteGAN}~\cite{li2019pastegan}& \textbf{SGDiff}~\cite{yang2022sgdiff}& \textbf{SceneGenie}~\cite{farshad2023scenegenie} &\textbf{SG-Adapter}\\ \midrule
        \textbf{FID}~$\downarrow$ & 99.1 & 79.1& 36.2& 62.4&\textbf{25.1}\\
        \textbf{Inception Score}~$\uparrow$ & 8.2& 12.3  & 17.8& 21.5&\textbf{57.8}\\ \bottomrule
    \end{tabular}
    }
    \label{tab:sg}
\end{table}

\begin{figure}[!h]
    \vspace{-0.2in}
    \centering
    \includegraphics[width=1.0\textwidth]{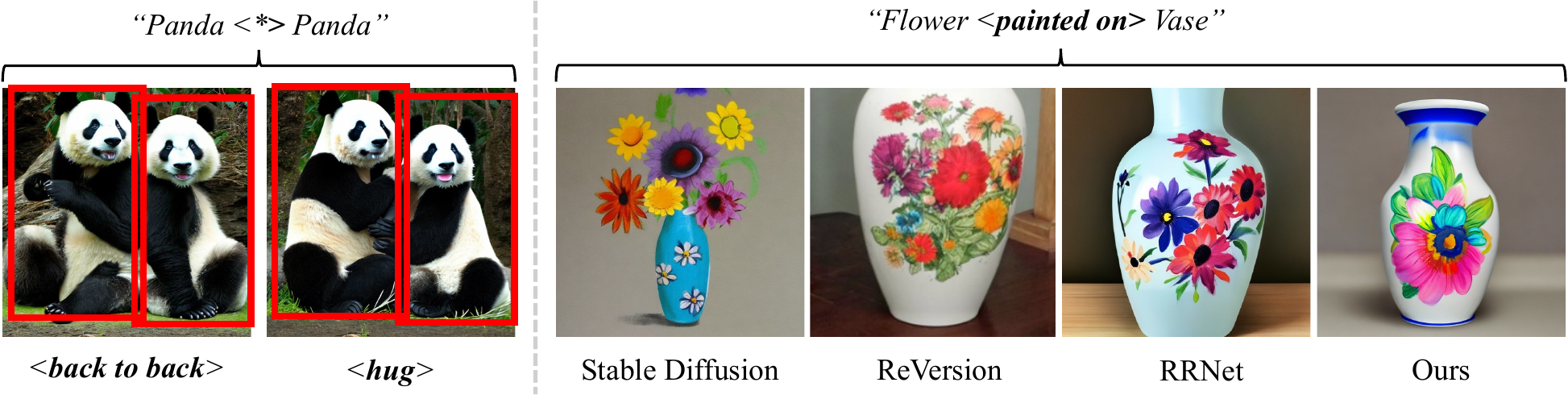} 
    \vspace{-0.2in}
    \caption{\textbf{Left:} The same layout appears visually different due to different relationships. \textbf{Right:} Our method is also capable of learning the customized single relationship.}
    \label{fig:relation_layout}
    \vspace{-0.2in}
\end{figure}

\section{Conclusion and Discussion}
\label{sec:conclusion}
This paper introduces the Scene Graph Adapter (SG-Adapter), a novel enhancement for text-to-image generation models. By integrating scene graph knowledge, the SG-Adapter significantly improves the contextual understanding of these models, ensuring images closely match their textual descriptions. The adapter employs an efficient triplet-token attention mechanism within a transformer architecture, allowing for more precise mapping of text to visual elements. To validate its effectiveness, the study utilized a specially curated dataset featuring multiple relations and high-quality annotations, showing the crucial role of clean, relation-rich data in multi-relational learning. As shown in Fig.\ref{fig:relation_layout}, SG-Adapter can also learn single complex relations as a by-product. Furthermore, we propose three effective metrics derived from GPT-4V\cite{achiam2023gpt} to evaluate the generated relation precisely. Lastly, our method could learn a new and complex single relation as a by-product, similar to ReVersion\cite{huang2023reversion} and RRNet\cite{wu2024relation}, as demonstrated in Fig.\ref{fig:relation_layout}.

As for limitation, to address data privacy concerns and follow the double-blind policy, we applied anonymization techniques to human faces in our MultiRels dataset, which may introduce some artifacts affecting image quality inevitably. We are actively exploring more sophisticated anonymization methods that preserve data integrity while ensuring privacy.

\small
\bibliography{main}

\begin{thebibliography}{36}
\providecommand{\natexlab}[1]{#1}
\providecommand{\url}[1]{\texttt{#1}}
\expandafter\ifx\csname urlstyle\endcsname\relax
  \providecommand{\doi}[1]{doi: #1}\else
  \providecommand{\doi}{doi: \begingroup \urlstyle{rm}\Url}\fi

\bibitem[Krishna et~al.(2017)Krishna, Zhu, Groth, Johnson, Hata, Kravitz, Chen, Kalantidis, Li, Shamma, et~al.]{krishna2017visual}
Ranjay Krishna, Yuke Zhu, Oliver Groth, Justin Johnson, Kenji Hata, Joshua Kravitz, Stephanie Chen, Yannis Kalantidis, Li-Jia Li, David~A Shamma, et~al.
\newblock Visual genome: Connecting language and vision using crowdsourced dense image annotations.
\newblock \emph{International journal of computer vision}, 123:\penalty0 32--73, 2017.

\bibitem[Achiam et~al.(2023)Achiam, Adler, Agarwal, Ahmad, Akkaya, Aleman, Almeida, Altenschmidt, Altman, Anadkat, et~al.]{achiam2023gpt}
Josh Achiam, Steven Adler, Sandhini Agarwal, Lama Ahmad, Ilge Akkaya, Florencia~Leoni Aleman, Diogo Almeida, Janko Altenschmidt, Sam Altman, Shyamal Anadkat, et~al.
\newblock Gpt-4 technical report.
\newblock \emph{arXiv preprint arXiv:2303.08774}, 2023.

\bibitem[Rombach et~al.(2022)Rombach, Blattmann, Lorenz, Esser, and Ommer]{rombach2022high}
Robin Rombach, Andreas Blattmann, Dominik Lorenz, Patrick Esser, and Bj{\"o}rn Ommer.
\newblock High-resolution image synthesis with latent diffusion models.
\newblock In \emph{Proceedings of the IEEE/CVF conference on computer vision and pattern recognition}, pages 10684--10695, 2022.

\bibitem[Saharia et~al.(2022)Saharia, Chan, Saxena, Li, Whang, Denton, Ghasemipour, Gontijo~Lopes, Karagol~Ayan, Salimans, et~al.]{saharia2022photorealistic}
Chitwan Saharia, William Chan, Saurabh Saxena, Lala Li, Jay Whang, Emily~L Denton, Kamyar Ghasemipour, Raphael Gontijo~Lopes, Burcu Karagol~Ayan, Tim Salimans, et~al.
\newblock Photorealistic text-to-image diffusion models with deep language understanding.
\newblock \emph{Advances in Neural Information Processing Systems}, 35:\penalty0 36479--36494, 2022.

\bibitem[Ramesh et~al.(2022)Ramesh, Dhariwal, Nichol, Chu, and Chen]{ramesh2022hierarchical}
Aditya Ramesh, Prafulla Dhariwal, Alex Nichol, Casey Chu, and Mark Chen.
\newblock Hierarchical text-conditional image generation with clip latents.
\newblock \emph{arXiv preprint arXiv:2204.06125}, 1\penalty0 (2):\penalty0 3, 2022.

\bibitem[Ho et~al.(2020)Ho, Jain, and Abbeel]{ho2020denoising}
Jonathan Ho, Ajay Jain, and Pieter Abbeel.
\newblock Denoising diffusion probabilistic models.
\newblock \emph{Advances in neural information processing systems}, 33:\penalty0 6840--6851, 2020.

\bibitem[Nichol et~al.(2021)Nichol, Dhariwal, Ramesh, Shyam, Mishkin, McGrew, Sutskever, and Chen]{nichol2021glide}
Alex Nichol, Prafulla Dhariwal, Aditya Ramesh, Pranav Shyam, Pamela Mishkin, Bob McGrew, Ilya Sutskever, and Mark Chen.
\newblock Glide: Towards photorealistic image generation and editing with text-guided diffusion models.
\newblock \emph{arXiv preprint arXiv:2112.10741}, 2021.

\bibitem[Dhariwal and Nichol(2021)]{dhariwal2021diffusion}
Prafulla Dhariwal and Alexander Nichol.
\newblock Diffusion models beat gans on image synthesis.
\newblock \emph{Advances in neural information processing systems}, 34:\penalty0 8780--8794, 2021.

\bibitem[Song et~al.(2020)Song, Meng, and Ermon]{song2020denoising}
Jiaming Song, Chenlin Meng, and Stefano Ermon.
\newblock Denoising diffusion implicit models.
\newblock \emph{arXiv preprint arXiv:2010.02502}, 2020.

\bibitem[Schuhmann et~al.(2022)Schuhmann, Beaumont, Vencu, Gordon, Wightman, Cherti, Coombes, Katta, Mullis, Wortsman, et~al.]{schuhmann2022laion}
Christoph Schuhmann, Romain Beaumont, Richard Vencu, Cade Gordon, Ross Wightman, Mehdi Cherti, Theo Coombes, Aarush Katta, Clayton Mullis, Mitchell Wortsman, et~al.
\newblock Laion-5b: An open large-scale dataset for training next generation image-text models.
\newblock \emph{Advances in Neural Information Processing Systems}, 35:\penalty0 25278--25294, 2022.

\bibitem[Lin et~al.(2014)Lin, Maire, Belongie, Hays, Perona, Ramanan, Doll{\'a}r, and Zitnick]{lin2014microsoft}
Tsung-Yi Lin, Michael Maire, Serge Belongie, James Hays, Pietro Perona, Deva Ramanan, Piotr Doll{\'a}r, and C~Lawrence Zitnick.
\newblock Microsoft coco: Common objects in context.
\newblock In \emph{Computer Vision--ECCV 2014: 13th European Conference, Zurich, Switzerland, September 6-12, 2014, Proceedings, Part V 13}, pages 740--755. Springer, 2014.

\bibitem[Radford et~al.(2021)Radford, Kim, Hallacy, Ramesh, Goh, Agarwal, Sastry, Askell, Mishkin, Clark, et~al.]{radford2021clip}
Alec Radford, Jong~Wook Kim, Chris Hallacy, Aditya Ramesh, Gabriel Goh, Sandhini Agarwal, Girish Sastry, Amanda Askell, Pamela Mishkin, Jack Clark, et~al.
\newblock Learning transferable visual models from natural language supervision.
\newblock In \emph{International conference on machine learning}, pages 8748--8763. PMLR, 2021.

\bibitem[Yang et~al.(2022{\natexlab{a}})Yang, Huang, Song, Hong, Li, Zhang, Cui, Ghanem, and Yang]{yang2022diffusion}
Ling Yang, Zhilin Huang, Yang Song, Shenda Hong, Guohao Li, Wentao Zhang, Bin Cui, Bernard Ghanem, and Ming-Hsuan Yang.
\newblock Diffusion-based scene graph to image generation with masked contrastive pre-training.
\newblock \emph{arXiv preprint arXiv:2211.11138}, 2022{\natexlab{a}}.

\bibitem[Johnson et~al.(2018)Johnson, Gupta, and Fei-Fei]{johnson2018sg2i}
Justin Johnson, Agrim Gupta, and Li~Fei-Fei.
\newblock Image generation from scene graphs.
\newblock In \emph{Proceedings of the IEEE conference on computer vision and pattern recognition}, pages 1219--1228, 2018.

\bibitem[Farshad et~al.(2023)Farshad, Yeganeh, Chi, Shen, Ommer, and Navab]{farshad2023scenegenie}
Azade Farshad, Yousef Yeganeh, Yu~Chi, Chengzhi Shen, B{\"o}jrn Ommer, and Nassir Navab.
\newblock Scenegenie: Scene graph guided diffusion models for image synthesis.
\newblock In \emph{Proceedings of the IEEE/CVF International Conference on Computer Vision}, pages 88--98, 2023.

\bibitem[Li et~al.(2019)Li, Ma, Bai, Duan, Wei, and Wang]{li2019pastegan}
Yikang Li, Tao Ma, Yeqi Bai, Nan Duan, Sining Wei, and Xiaogang Wang.
\newblock Pastegan: A semi-parametric method to generate image from scene graph.
\newblock \emph{Advances in Neural Information Processing Systems}, 32, 2019.

\bibitem[Tang et~al.(2020)Tang, Niu, Huang, Shi, and Zhang]{tang2020unbiased}
Kaihua Tang, Yulei Niu, Jianqiang Huang, Jiaxin Shi, and Hanwang Zhang.
\newblock Unbiased scene graph generation from biased training.
\newblock In \emph{Proceedings of the IEEE/CVF conference on computer vision and pattern recognition}, pages 3716--3725, 2020.

\bibitem[Li et~al.(2023{\natexlab{a}})Li, Liu, Wu, Mu, Yang, Gao, Li, and Lee]{li2023gligen}
Yuheng Li, Haotian Liu, Qingyang Wu, Fangzhou Mu, Jianwei Yang, Jianfeng Gao, Chunyuan Li, and Yong~Jae Lee.
\newblock Gligen: Open-set grounded text-to-image generation.
\newblock In \emph{Proceedings of the IEEE/CVF Conference on Computer Vision and Pattern Recognition}, pages 22511--22521, 2023{\natexlab{a}}.

\bibitem[Zhang et~al.(2023)Zhang, Rao, and Agrawala]{zhang2023controlnet}
Lvmin Zhang, Anyi Rao, and Maneesh Agrawala.
\newblock Adding conditional control to text-to-image diffusion models.
\newblock In \emph{Proceedings of the IEEE/CVF International Conference on Computer Vision}, pages 3836--3847, 2023.

\bibitem[Mou et~al.(2023)Mou, Wang, Xie, Zhang, Qi, Shan, and Qie]{mou2023t2i}
Chong Mou, Xintao Wang, Liangbin Xie, Jian Zhang, Zhongang Qi, Ying Shan, and Xiaohu Qie.
\newblock T2i-adapter: Learning adapters to dig out more controllable ability for text-to-image diffusion models.
\newblock \emph{arXiv preprint arXiv:2302.08453}, 2023.

\bibitem[Huang et~al.(2023)Huang, Wu, Jiang, Chan, and Liu]{huang2023reversion}
Ziqi Huang, Tianxing Wu, Yuming Jiang, Kelvin~CK Chan, and Ziwei Liu.
\newblock Reversion: Diffusion-based relation inversion from images.
\newblock \emph{arXiv preprint arXiv:2303.13495}, 2023.

\bibitem[Feng et~al.(2022)Feng, He, Fu, Jampani, Akula, Narayana, Basu, Wang, and Wang]{feng2022ssd}
Weixi Feng, Xuehai He, Tsu-Jui Fu, Varun Jampani, Arjun Akula, Pradyumna Narayana, Sugato Basu, Xin~Eric Wang, and William~Yang Wang.
\newblock Training-free structured diffusion guidance for compositional text-to-image synthesis.
\newblock \emph{arXiv preprint arXiv:2212.05032}, 2022.

\bibitem[Li et~al.(2023{\natexlab{b}})Li, Keuper, Zhang, and Khoreva]{li2023divide}
Yumeng Li, Margret Keuper, Dan Zhang, and Anna Khoreva.
\newblock Divide \& bind your attention for improved generative semantic nursing.
\newblock \emph{arXiv preprint arXiv:2307.10864}, 2023{\natexlab{b}}.

\bibitem[Zheng et~al.(2023)Zheng, Zhou, Li, Qi, Shan, and Li]{zheng2023layoutdiffusion}
Guangcong Zheng, Xianpan Zhou, Xuewei Li, Zhongang Qi, Ying Shan, and Xi~Li.
\newblock Layoutdiffusion: Controllable diffusion model for layout-to-image generation.
\newblock In \emph{Proceedings of the IEEE/CVF Conference on Computer Vision and Pattern Recognition}, pages 22490--22499, 2023.

\bibitem[Chen et~al.(2023)Chen, Laina, and Vedaldi]{chen2023trainingfreelayout}
Minghao Chen, Iro Laina, and Andrea Vedaldi.
\newblock Training-free layout control with cross-attention guidance.
\newblock \emph{arXiv preprint arXiv:2304.03373}, 2023.

\bibitem[Couairon et~al.(2023)Couairon, Careil, Cord, Lathuili{\`e}re, and Verbeek]{couairon2023zerolayout}
Guillaume Couairon, Marl{\`e}ne Careil, Matthieu Cord, St{\'e}phane Lathuili{\`e}re, and Jakob Verbeek.
\newblock Zero-shot spatial layout conditioning for text-to-image diffusion models.
\newblock In \emph{Proceedings of the IEEE/CVF International Conference on Computer Vision}, pages 2174--2183, 2023.

\bibitem[Yang et~al.(2022{\natexlab{b}})Yang, Huang, Song, Hong, Li, Zhang, Cui, Ghanem, and Yang]{yang2022sgdiff}
Ling Yang, Zhilin Huang, Yang Song, Shenda Hong, Guohao Li, Wentao Zhang, Bin Cui, Bernard Ghanem, and Ming-Hsuan Yang.
\newblock Diffusion-based scene graph to image generation with masked contrastive pre-training.
\newblock \emph{arXiv preprint arXiv:2211.11138}, 2022{\natexlab{b}}.

\bibitem[Wu et~al.(2023)Wu, Wei, and Lin]{wu2023sggan}
Yang Wu, Pengxu Wei, and Liang Lin.
\newblock Scene graph to image synthesis via knowledge consensus.
\newblock In \emph{Proceedings of the AAAI Conference on Artificial Intelligence}, volume~37, pages 2856--2865, 2023.

\bibitem[Wu et~al.(2024)Wu, Yang, and Wang]{wu2024relation}
Yinwei Wu, Xingyi Yang, and Xinchao Wang.
\newblock Relation rectification in diffusion model.
\newblock \emph{arXiv preprint arXiv:2403.20249}, 2024.

\bibitem[Hu et~al.(2021)Hu, Shen, Wallis, Allen-Zhu, Li, Wang, Wang, and Chen]{hu2021lora}
Edward~J Hu, Yelong Shen, Phillip Wallis, Zeyuan Allen-Zhu, Yuanzhi Li, Shean Wang, Lu~Wang, and Weizhu Chen.
\newblock Lora: Low-rank adaptation of large language models.
\newblock \emph{arXiv preprint arXiv:2106.09685}, 2021.

\bibitem[Heusel et~al.(2018)Heusel, Ramsauer, Unterthiner, Nessler, and Hochreiter]{heusel2018gans}
Martin Heusel, Hubert Ramsauer, Thomas Unterthiner, Bernhard Nessler, and Sepp Hochreiter.
\newblock Gans trained by a two time-scale update rule converge to a local nash equilibrium, 2018.

\bibitem[Caesar et~al.(2018)Caesar, Uijlings, and Ferrari]{caesar2018coco}
Holger Caesar, Jasper Uijlings, and Vittorio Ferrari.
\newblock Coco-stuff: Thing and stuff classes in context.
\newblock In \emph{Proceedings of the IEEE conference on computer vision and pattern recognition}, pages 1209--1218, 2018.

\bibitem[Wang et~al.(2023)Wang, Farnia, Lin, Shen, and Yu]{wang2023evaluation}
Zixiao Wang, Farzan Farnia, Zhenghao Lin, Yunheng Shen, and Bei Yu.
\newblock On the evaluation of generative models in distributed learning tasks.
\newblock \emph{arXiv preprint arXiv:2310.11714}, 2023.

\bibitem[Ruiz et~al.(2023)Ruiz, Li, Jampani, Pritch, Rubinstein, and Aberman]{ruiz2023dreambooth}
Nataniel Ruiz, Yuanzhen Li, Varun Jampani, Yael Pritch, Michael Rubinstein, and Kfir Aberman.
\newblock Dreambooth: Fine tuning text-to-image diffusion models for subject-driven generation.
\newblock In \emph{Proceedings of the IEEE/CVF Conference on Computer Vision and Pattern Recognition}, pages 22500--22510, 2023.

\bibitem[Plummer et~al.(2015)Plummer, Wang, Cervantes, Caicedo, Hockenmaier, and Lazebnik]{plummer2015flickr30k}
Bryan~A Plummer, Liwei Wang, Chris~M Cervantes, Juan~C Caicedo, Julia Hockenmaier, and Svetlana Lazebnik.
\newblock Flickr30k entities: Collecting region-to-phrase correspondences for richer image-to-sentence models.
\newblock In \emph{Proceedings of the IEEE international conference on computer vision}, pages 2641--2649, 2015.

\bibitem[Salimans et~al.(2016)Salimans, Goodfellow, Zaremba, Cheung, Radford, and Chen]{salimans2016improved}
Tim Salimans, Ian Goodfellow, Wojciech Zaremba, Vicki Cheung, Alec Radford, and Xi~Chen.
\newblock Improved techniques for training gans.
\newblock \emph{Advances in neural information processing systems}, 29, 2016.

\end{thebibliography}
\bibliographystyle{unsrtnat}






\newpage
\appendix

\section{Appendix}
\subsection{Initial Experiment with Scene Graph Attention Mask}
\label{subsec:init_attempt}

\begin{figure}[htbp]
    \centering
    \includegraphics[width=1.0\linewidth]{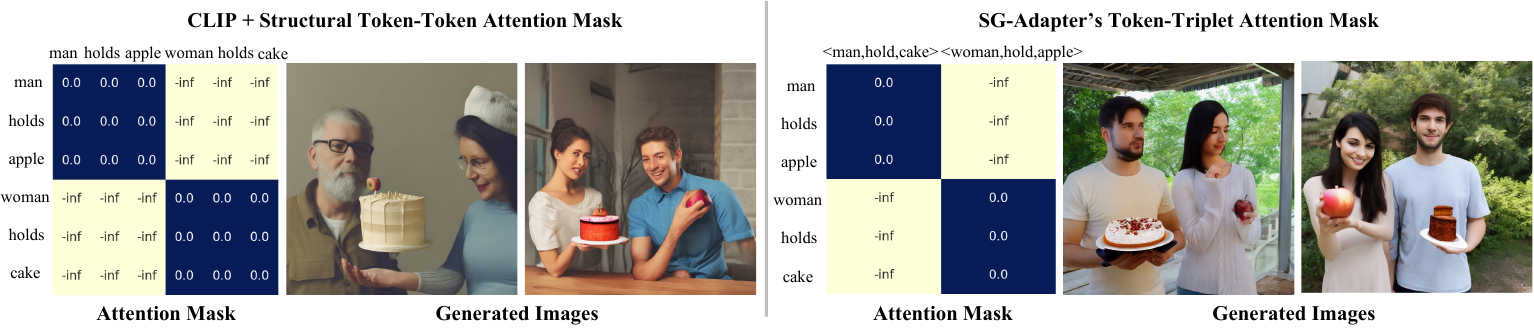}
    \caption{\textbf{Results of Initial Attempt and SG-Adapter.} Directly integrating the structural token-token attention mask in a hard way within the CLIP model could break the delicate balance of learned sequential dependencies and fail to ensure accurate semantic structure in the generated images. Instead, our SG-Adapter makes use of a novel token-triplet attention mask in a learnable way to correct the unreasonable token-token interactions and guarantee precise relation correspondence.}
    \label{fig-init-attempt}
\end{figure}

Our initial attempts involved the utilization of an adjusted attention mask \( \mathbf{M}^{\tau} \) within the transformer's attention mechanism of the CLIP text encoder, aiming to ensure intra-triplet attention cohesion and maintain the semantic structure of caption. 
This integration, however, did not translate to empirical improvements as shown in Fig.~\ref{fig-init-attempt}. 
We observed that directly manipulating the attention mask of the pre-trained CLIP model could disrupt the delicate balance of learned sequential dependencies, adversely affecting the quality of text-to-image synthesis due to inconsistencies between the model's training and inference methodologies.

\subsection{MultiRels Benchmark Details}
\label{sec:MultiRels dataset}

As mentioned in the main paper, we organized the MultiRels into two parts: Reversion and Multiple Relations. This section will introduce more detailed information about the Multiple Relations part.

The Multiple Relations part contains 210 samples. We initially plan to collect all the images from the Internet and then label them manually. However, after we had collected dozens of images, we found this way not efficient enough since there are very few images that contain multiple clear and salient relations. Besides, a text describing a multiple relations image tends to be long so the relevance of retrieved images will also decrease with the longer text. Therefore we only collect 40 multi-relations images from the Internet by retrieving a long text.

\begin{figure*}[t]
    \centering
    \includegraphics[width=1.0\linewidth]{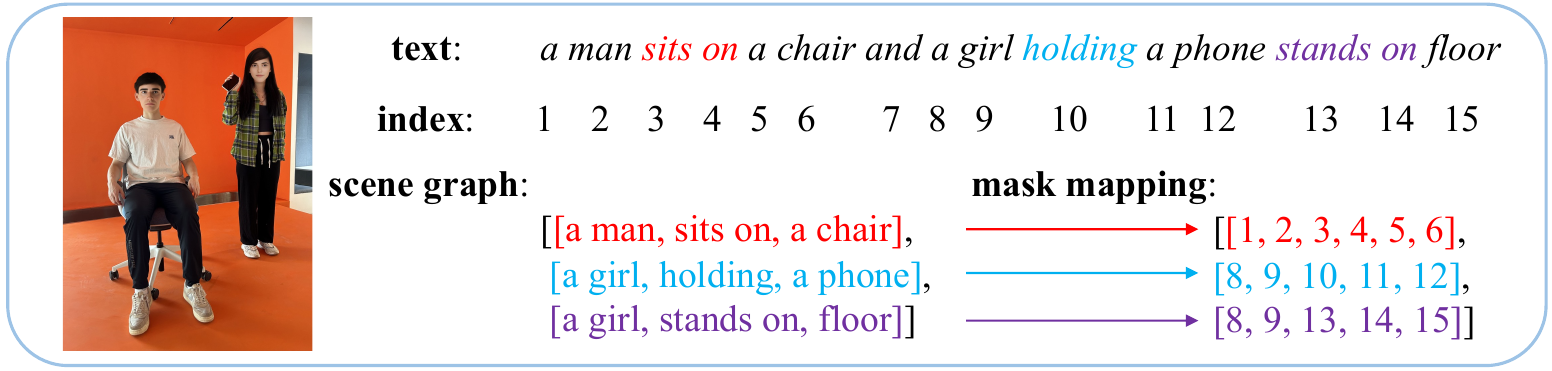}
    \caption{\textbf{Example 1 of MultiRels}}
    \label{fig-multiRels1}
    \vspace{0.05in}
\end{figure*}

\begin{figure*}[t]
    \centering
    \includegraphics[width=1.0\linewidth]{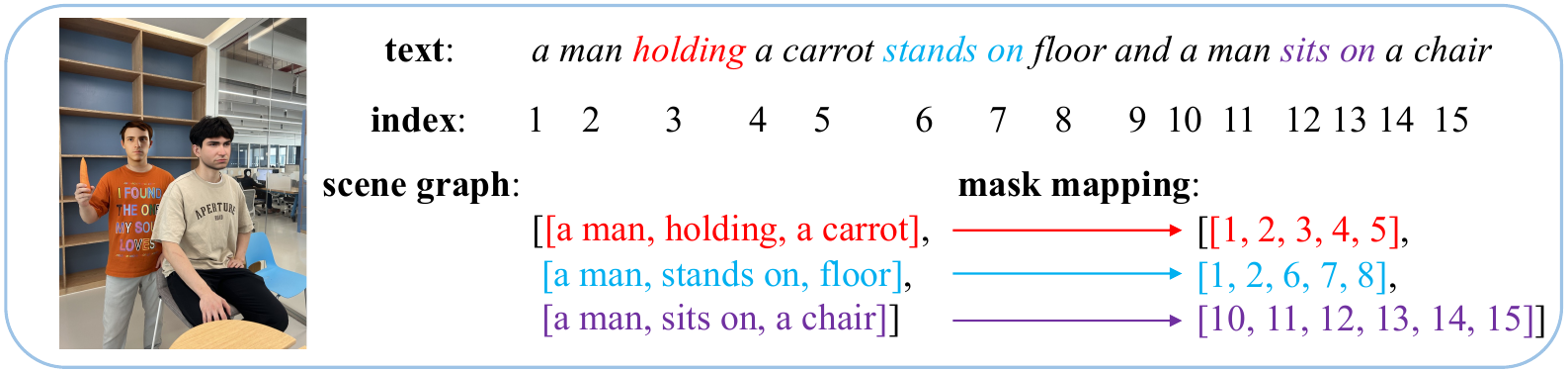}
    \caption{\textbf{Example 2 of MultiRels}}
    \label{fig-multiRels2}
    \vspace{0.05in}
\end{figure*}

\begin{figure*}[t]
    \centering
    \includegraphics[width=1.0\linewidth]{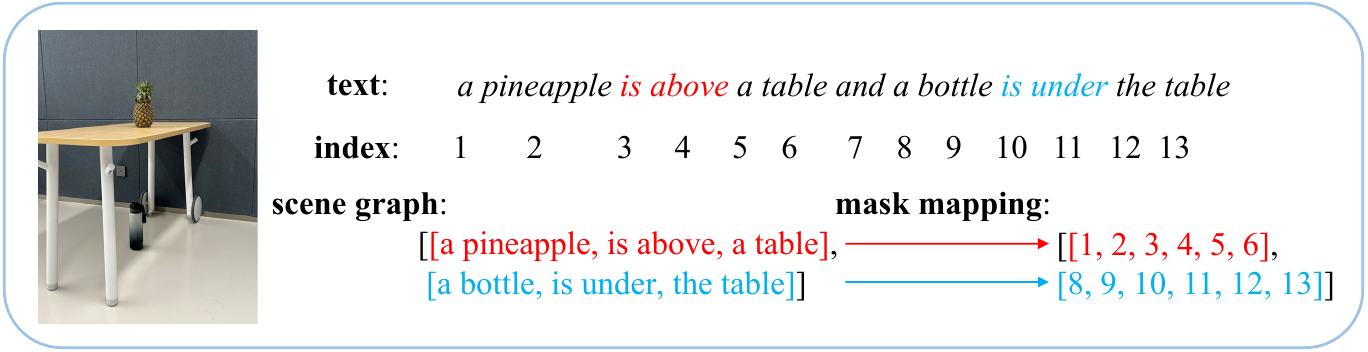}
    \caption{\textbf{Example 3 of MultiRels}}
    \label{fig-multiRels3}
    \vspace{0.05in}
\end{figure*}

\begin{figure*}[t]
    \centering
    \includegraphics[width=1.0\linewidth]{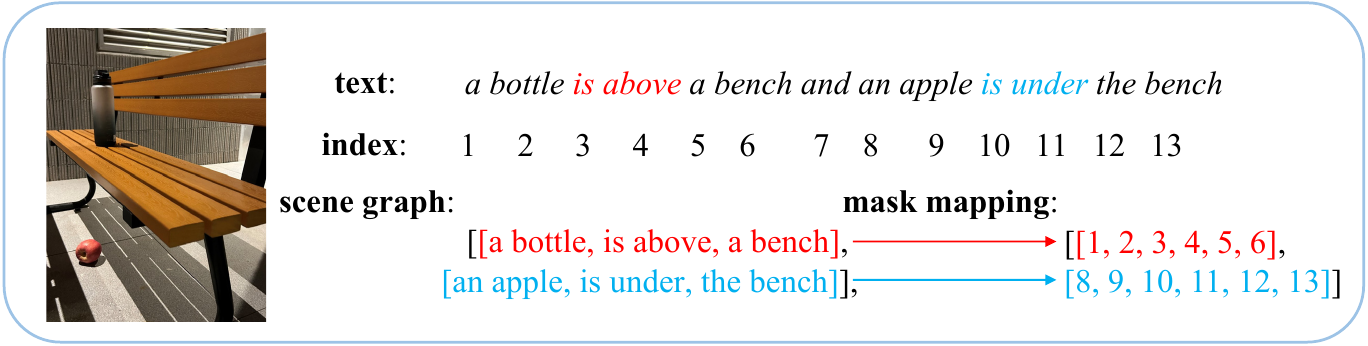}
    \caption{\textbf{Example 4 of MultiRels}}
    \label{fig-multiRels4}
    \vspace{0.05in}
\end{figure*}

On the contrary, 170 images in the Multiple Relations part were taken by ourselves. 
\begin{enumerate}
    \item Concerning human relations, we mainly focus on ``\textit{holding}", ``\textit{drinking}", ``\textit{stands on}", ``\textit{sits on}" such kinds of simple human actions, combining random 2-3 relations from them to result in 15 different text templates containing multiple relations finally. To make the collected images as diverse as possible, we arranged for 5 volunteers to participate in the photoshoot in 6 different environmental backgrounds both indoor and outdoor. \textbf{We re-paint faces in these photos locally for the privacy of the volunteers and the blind paper review}. We examples of the above data in Fig.\ref{fig-multiRels1}, Fig.\ref{fig-multiRels2}.
    \item For object relations, we consider the positional relationships like ``\textit{is above}", and ``\textit{is under}". We place various normal objects, e.g. fruits and daily necessities, on the same/different side of an object like a table, chair, bench, and so on. Also, for the diversity of the data set, we take photos of these objects in 6 different indoor and outdoor environments. There are examples provided in Fig.\ref{fig-multiRels3}, Fig.\ref{fig-multiRels4}.
\end{enumerate}
The 15 templates we adopt in human relations are as follows:
\begin{enumerate}
    \item \textit{a man stands on floor and a woman sits on a chair}.
    \item \textit{a woman stands on floor and a man sits on a chair}.
    \item \textit{a man stands on floor and a woman stands on floor}.
    \item \textit{a man sits on a chair and a woman sits on a chair}.
    \item \textit{a man drinking milk and a woman drinking cola}.
    \item \textit{a man sits on a chair and a woman holding a \(\langle obj \rangle \) stands on floor}.
    \item \textit{a woman sits on a chair and a man holding a \(\langle obj \rangle \) stands on floor}.
    \item \textit{a man drinking juice sits on a chair and a woman stands on floor}.
    \item \textit{a man holding a \(\langle obj \rangle \) stands on floor and a woman drinking water stands on floor}.
    \item \textit{a man stands on floor and a man sits on a chair}.
    \item \textit{a man drinking water sits on a chair and a man holding a \(\langle obj \rangle \) stands on floor}.
    \item \textit{a man holding a \(\langle obj \rangle \) stands on floor and a man sits on a chair}.
    \item \textit{a man holding a \(\langle obj \rangle \) stands on floor and a man stands on floor}.
    \item \textit{a man drinking juice and a man drinking water}.
    \item \textit{a man drinking water and a man drinking water}.
\end{enumerate}

To see the complete data set and its corresponding metadata, please refer to the compressed file MultiRels.zip.

\subsection{Additional Qualitative Results}
\begin{figure*}[ht]
    \centering
    \includegraphics[width=1.0\linewidth]{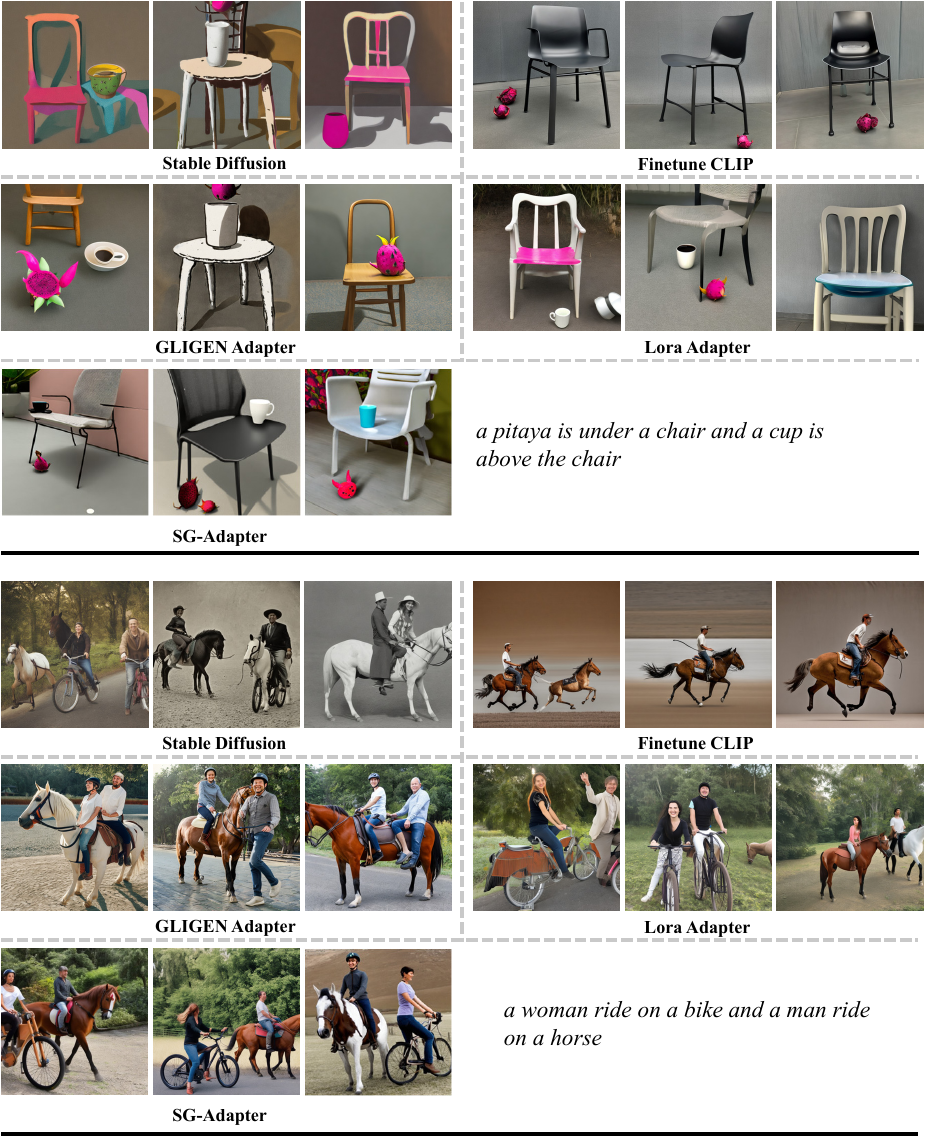}
    \vspace{-0.2in}
    \caption{\textbf{More qualitative results-1.}}
    \label{fig-user qulitative_results_1}
\end{figure*}

\begin{figure*}[ht]
    \centering
    \includegraphics[width=1.0\linewidth]{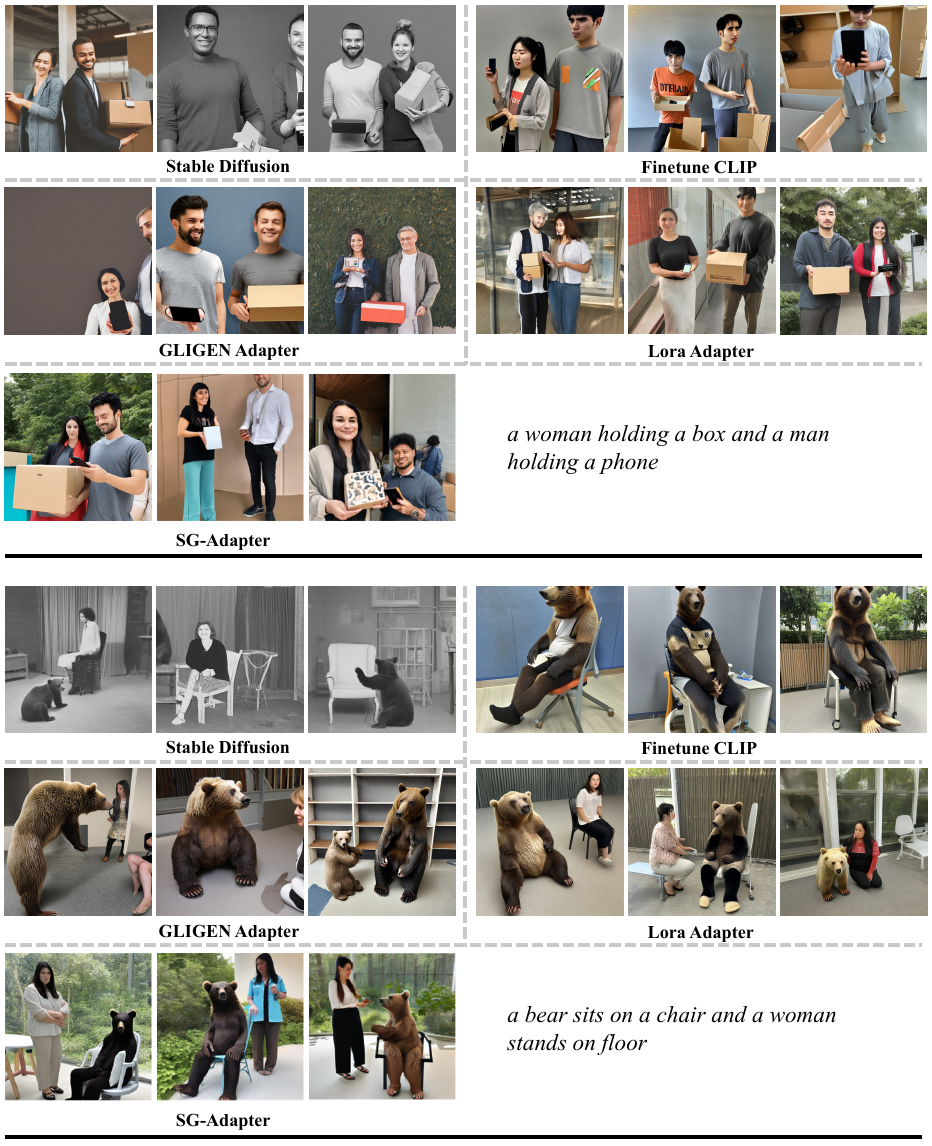}
    \vspace{-0.2in}
    \caption{\textbf{More qualitative results-2.}}
    \label{fig-user qulitative_results_2}
\end{figure*}

We present additional qualitative results of our SG-Adapter compared with other baseline methods in Fig.\ref{fig-user qulitative_results_1}, Fig.\ref{fig-user qulitative_results_2}.

\subsection{GPT-4V Prompts}
\label{subsec:prompt}
\noindent \textbf{Prompt to Extract Scene Graph from Image:} \textit{Please extract the scene graph of the given image. The scene graph just needs to include the relations of the salient objects and exclude the background. The scene graph should be a list of triplets like ["subject", "predicate", "object"]. \\
Both subject and object should be selected from the following list: ["an astronaut", "a ball", "a cake", "a box", "a television", "a table", "a horse", "a pitaya", "a woman", "a book", "a laptop", "a bottle", "a banana", "a sofa", "floor", "water", "an apple", "a chair", "a pineapple", "an umbrella", "a boy", "a paper", "a bear", "a girl", "a panda", "a cup", "a man", "a bike", "a carrot", "a phone"].\\ 
The predicate should be selected from the following list: ["stands on", "is above", "drinking", "is under", "sits back to back with", "ride on", "holding", "sits on"]\\
Besides the scene graph, please also output the objects list in the image like ["object1", "object2", ..., "object"]. The object should be also selected from the above-mentioned object list. The output should only contain the scene graph and the object list.}

\noindent \textbf{Prompt to Parse Caption to Scene Graph:}\textit{Here I have a group of captions and please help me to parse each one. Each caption should be transformed to a Scene Graph reasonably which is composed of some relations. A relation is a triplet like [subject, predicate, object] and please replace the original pronouns with reasonable nouns. Both subject and object should only have one object or person. "and" relation should not be included in the Scene Graph. Besides, I want to get the indexes of all of the subjects, predicates, and objects in the original caption, which is called mapping here. \\ 
For example, the caption is: a boy holding a bottle shakes hands with a girl sitting on a bench.\\
The corresponding Scene Graph should be: [[a boy, holding, a bottle], [a boy, shakes hands with, a girl], [a girl, sitting on, a bench]]. \\
The indexes of each word(punctuation is also considered a word) in the caption(called all\_words\_indexes) are:{"a":0, "boy":1, "holding":2, "a" :3, "bottle":4, "shakes":5, "hands":6, "with":7, "a":8, "girl":9, "sitting":10, "on":11, "a":12, "bench":13}.\\
The index of every word in the Scene Graph(i.e., the mapping) should be:[{"a":0, "boy":1, "holding":2," a":3, "bottle":4},{"a":0, "boy":1, "shakes":5, "hands":6, "with":7}, {"a":8, "girl":9, "sitting":10, "on": 11,"a":12, "bench":13}]\\
The indexes of all of the subject, predicate, and object in the original caption should be highly precise. The results of each caption are data like:\\
scene graph:\\
all\_words\_indexes:\\
mapping:\\
When generating the mapping please refer to the scene graph and the all\_words\_indexes to ensure the correct result.\\
The captions are:}

\subsection{Implementation Details}
All the experiments are conducted on \(768\times768\) image resolution and all the models are trained on a single A100 GPU for several hours. We adopt SD 2.1 as our base model. We set training batch size 4, learning rate 1e-5. We adopt the optimizer AdamW and most models converge around 12000 to 14000 iterations. During the sampling process, we use a parameter \(\tau\) to balance the ability to control and the image quality. For a diffusion process with \(T\) steps, we could use scene graph guided inference at the first \(\tau*T\) steps and use the standard inference for the remaining \((1-\tau)*T\) steps. In this paper, \(\tau\) set to 0.3 is enough to achieve relation control while maintaining the image quality.

\clearpage

\clearpage


\newpage

\end{document}